\documentclass[letterpaper]{article} 
\usepackage[]{aaai25}
\usepackage{times}  
\usepackage{helvet}  
\usepackage{courier}  
\usepackage[hyphens]{url}  
\usepackage{graphicx} 
\usepackage{natbib}  
\usepackage{caption} 
\usepackage{algorithm}
\usepackage{algorithmic}

\usepackage{newfloat}
\usepackage{listings}
\DeclareCaptionStyle{ruled}{labelfont=normalfont,labelsep=colon,strut=off} 
\floatstyle{ruled}
\newfloat{listing}{tb}{lst}{}
\floatname{listing}{Listing}
\usepackage{xcolor}
\usepackage{amsmath}
\usepackage{amsfonts}
\usepackage{booktabs}
\usepackage{multirow}

\title{Hierarchical Mixture of Experts: Generalizable Learning for High-Level Synthesis}
\author{
    Weikai Li, Ding Wang, Zijian Ding, Atefeh Sohrabizadeh, Zongyue Qin, \\ Jason Cong, Yizhou Sun
}
\affiliations{
    University of California, Los Angeles \\
    weikaili@cs.ucla.edu, allenwang2333@gmail.com, \{bradyd, atefehsz, qinzongyue, cong, yzsun\}@cs.ucla.edu
}

\begin{document}

\maketitle

\begin{abstract}
High-level synthesis (HLS) is a widely used tool in designing Field Programmable Gate Array (FPGA). HLS enables FPGA design with software programming languages by compiling the source code into an FPGA circuit. The source code includes a program (called ``kernel'') and several pragmas that instruct hardware synthesis, such as parallelization, pipeline, etc. While it is relatively easy for software developers to design the program, it heavily relies on hardware knowledge to design the pragmas, posing a big challenge for software developers. Recently, different machine learning algorithms, such as GNNs, have been proposed to automate the pragma design via performance prediction. However, when applying the trained model on new kernels, the significant domain shift often leads to unsatisfactory performance. We propose a more domain-generalizable model structure: a two-level hierarchical Mixture of Experts (MoE), that can be flexibly adapted to any GNN model. Different expert networks can learn to deal with different regions in the representation space, and they can utilize similar patterns between the old kernels and new kernels. In the low-level MoE, we apply MoE on three natural granularities of a program: node, basic block, and graph. The high-level MoE learns to aggregate the three granularities for the final decision. To train the hierarchical MoE stably, we further propose a two-stage training method to avoid expert polarization. Extensive experiments verify the effectiveness of the proposed hierarchical MoE. We publicized our codes at https://github.com/weikai-li/HierarchicalMoE.
\end{abstract}


\section{Introduction}

With the heated demand for domain-specific accelerators, field-programmable gate arrays (FPGAs) are widely used. It is, however, labor-intensive to write register-transfer-level hardware description languages, such as VHDL and Verilog. High-level synthesis (HLS) provides a much easier solution by compiling a source code written in C++ into an FPGA circuit~\cite{HLS_background1,HLS_background2,HLS_background3}.

\begin{figure*}[tbp]
\includegraphics[width=\linewidth]{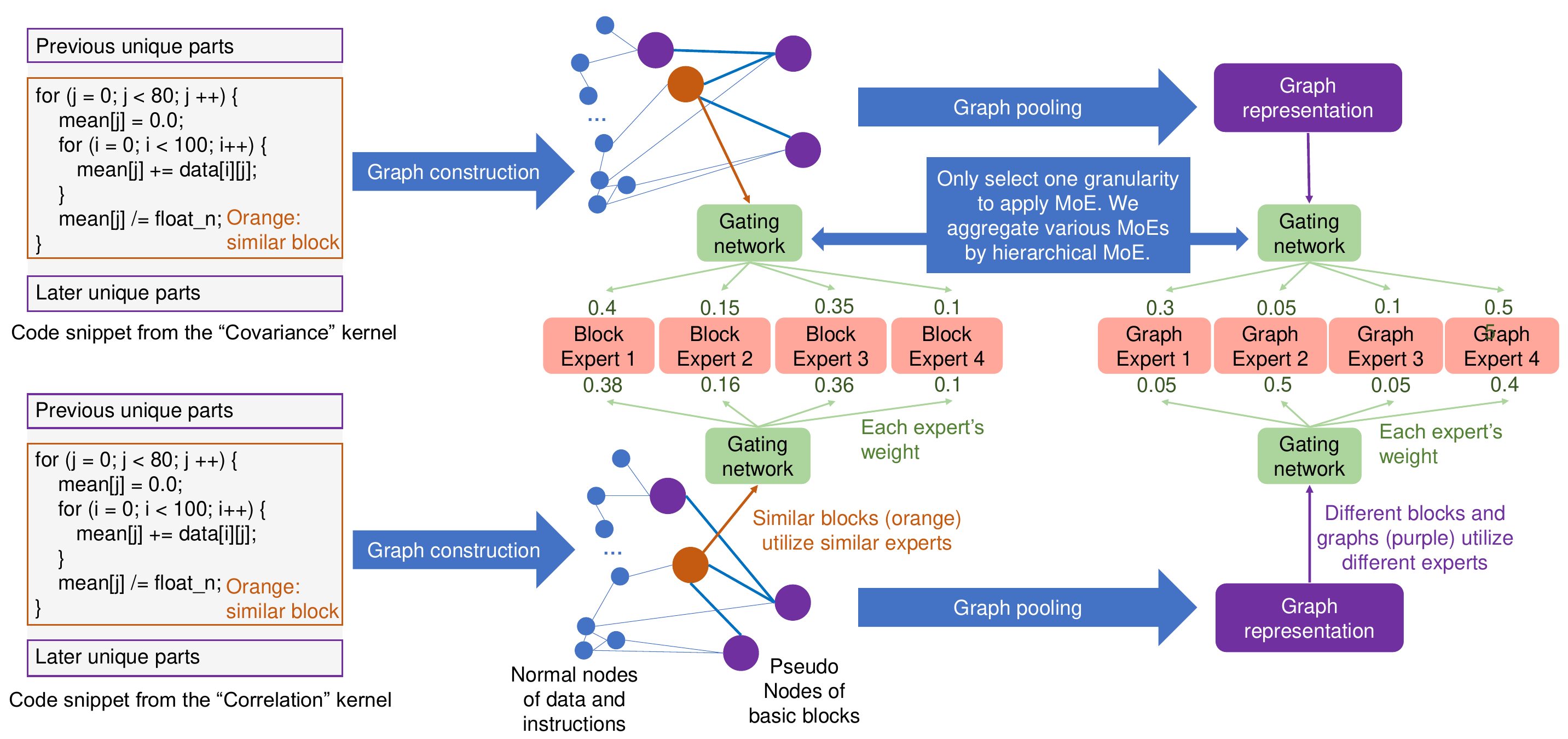}
\caption{Motivation of utilizing MoE. The two kernels share a similar nested loop, while the other parts are different. Similar parts can utilize similar experts via similar gating, while different parts can utilize different experts. This is just an illustration. Actually, we do not apply MoE on different granularities in the same model in this way. We aggregate them like Figure~\ref{fig:model}.
}
\label{fig:intro_challenge}
\vspace{-2mm}
\end{figure*}

The source code consists of a C++ program (also called a ``kernel'') that describes the FPGA's functional behaviors and several pragmas inserted into the program. The pragmas instruct the hardware synthesis process, such as parallelization, pipeline, and loop tiling, as illustrated in the code snippets in Fig.~\ref{fig:intro_challenge}. Different pragmas can lead to significantly different FPGA performances, so it is very important fo find the most effective pragmas. It still requires a great deal of hardware knowledge to design the pragmas, posing a great challenge for software developers~\cite{gnn-dse}.

Recent works have designed several methods to automate pragma design for a given C++ kernel. Heuristic methods use bottleneck analysis~\cite{AutoDSE} or non-linear programming based on lower-bound estimation~\cite{NLP-DSE,Sisyphus}. Machine learning methods train a surrogate model to predict the FPGA's performance from the source code, so that we can rely on the model prediction to find the best pragmas without running time-consuming HLS. HLS data is scarce, so it is difficult to train a large language model, and graph neural networks (GNNs) based on the control data flow graph are widely used~\cite{gnn-dse,gnn-dse-maml,harp,gnn_hls_model1,gnn_hls_model2,gnn_hls_model3}. A recent work~\cite{progsg} also uses both GNN and lightweight language models.

While machine learning models have better learning abilities than heuristic methods, they suffer from poor generalizability. A trained model usually fails to perform well on new kernels. This can be viewed as a domain generalization problem~\cite{gnn-dse-maml,hls_mlp_transfer}, where each kernel is one domain. The first challenge is that the domain difference is too big, and the difference is informative for the prediction. Previous domain generalization methods for GNN usually employ adversarial training to align the representation space between different domains~\cite{AdaGCN,DANE,UDA-GCN,Domain_adatpive_GNN2023}, or do data augmentation to learn invariant representation under risk extrapolation~\cite{EERM,FLOOD}. They are not directly applicable to our task, since we do not want to force different domains' representations to fit in the same distribution. The second challenge is that it is time-consuming to run HLS to acquire labels on new kernels, with each run taking minutes to hours~\cite{harp}.

While the HLS prediction task is challenging to domain generalization, it also provides a unique opportunity: different kernels usually share some similar substructures. As shown in Fig~\ref{fig:intro_challenge}, it would be beneficial if we could utilize the knowledge learned from similar substructures on the old kernels to apply on the new kernels. MoE~\cite{jacobs1991,shazeer2017} is an appropriate solution. It computes the weighted sum of the outputs from several expert networks instead of a single network. Different experts can specialize in different regions in the hidden representation space, so similar substructures in different domains can be processed by similar experts.
GraphMETRO~\cite{GraphMETRO} has demonstrated MoE's effect on improving GNNs' generalizability. It defines several types of domain shifts and trains each expert to deal with one type by data augmentation. We take an orthogonal direction, eliminate the need to define the domain shifts, and focus on the model structure rather than data augmentation.

The HLS graph naturally has three levels of granularity: nodes, blocks, and the whole graph. To build an MoE model suitable for this specific task, we consider constructing MoE on these three granularities. We find that different kernels benefit from applying MoE on different granularities. A natural question is which granularity to use. Simply applying MoE on all three granularities simultaneously does not work well. Therefore, we propose a higher-level MoE to aggregate the three low-level MoEs according to the need, which forms a two-level hierarchical MoE structure. It can flexibly adjust the weight of each low-level MoE based on the data. To the best of our knowledge, while some papers~\cite{shazeer2017} have studied hierarchical MoE to improve memory efficiency, this is the first study to demonstrate the performance gain of hierarchical MoE.

Nonetheless, optimization with the hierarchical MoE is challenging. Different from previous MoE papers where all the expert networks have the same or similar structures, our three low-level MoEs have very different structures, and thus very different convergence speeds. As a result, training easily leads to expert polarization and mode collapse. It leads to only using the single expert that converges the fastest. Thus, we propose a two-stage training strategy to stabilize the training. We conduct extensive experiments on the benchmark dataset, HLSyn~\cite{hlsyn}, and the experiment results reveal the effectiveness of hierarchical MoE. In summary, we make the following contributions:
\begin{itemize}
    \item We identify the difficulties of domain generalizable learning in HLS prediction and formalize this problem.

    \item We propose hierarchical MoE to address these challenges and a two-stage training approach to stabilize its training.

    \item Extensive experiments verify our methods' effectiveness, where the geometric mean speedup of the FPGA design is 26.6\% higher than that of the baseline method.
\end{itemize}

\section{Related Work}

\subsection{Machine Learning for HLS Prediction}
There are two directions for HLS pragma design automation. The first direction is heuristic-driven. AutoDSE~\cite{AutoDSE} uses bottleneck-guided searching based on the HLS feedback, and it is well-generalizable. Another heuristic method is non-linear programming, which performs well on affine kernels based on a lower-bound objective function~\cite{NLP-DSE,Sisyphus}. The second direction is data-driven, which trains a surrogate machine learning model to predict the FPGA's latency and resource utilization. GNNs are widely used~\cite{gnn-dse,harp,gnn_hls_model1,gnn_hls_model3,gnn_hls_model2,gnn_hls_model4,progsg}.

While machine learning models perform better than heuristic methods with sufficient training data, they suffer from poor generalizability. Some previous works have tried to solve this issue. \cite{hls_map_transfer} transforms the best pragma design from the most similar source kernel to the target kernel, but its transformation is too simple and limited. \cite{hls_mlp_transfer} trains a separate adaptor for each domain, while \cite{gnn-dse-maml} uses the meta-learning method, MAML~\cite{MAML}, to find generalizable parameter initializations. However, we have many kernels, so it is difficult to train MAML to get stable results. We only have hundreds of data points per kernel, so it is difficult to train an adaptor. We explore an orthogonal direction: to improve the model structure's generalizability.

\subsection{Domain Generalization for Graph Neural Network}
Domain generalization aims to reduce the performance gap between the source and target domains. Most related works~\cite{AdaGCN,DANE,UDA-GCN,Domain_adatpive_GNN2023} employ adversarial training to align the representation space between the source and target domains. Some papers~\cite{EERM,FLOOD} design specific data augmentation methods and learn invariant representations under risk extrapolation. However, we cannot directly align the representation space of different kernels since their difference is large and is useful for the prediction. GraphMETRO~\cite{GraphMETRO} predefines several types of domain shifts and uses MoE to deal with them, where each expert deals with one type. We take an orthogonal direction and eliminate the need for hardware knowledge to define the types of domain shifts, and our model based on the three granularities is specifically designed for this task.

\subsection{Mixture of Experts (MoE)}
MoE uses several expert networks and calculates the weighted sum of their outputs~\cite{jacobs1991,shazeer2017}. It is useful in domain transfer learning in computer vision~\cite{moe_transfer_cv1,moe_transfer_cv2}. It has also been utilized in GNN to improve the performance~\cite{gmoe,node_moe_theory}, diversify node representations in fairness-augmented graphs~\cite{GFAME}, and deal with class imbalance~\cite{GraphDIVE}. Previous work~\cite{shazeer2017} has employed hierarchical expert routing to improve memory efficiency, enabling a larger number of experts. However, previous papers have not shown the performance gain of hierarchical MoEs over regular MoEs when using the same number of total experts.

\section{Preliminary}

\subsection{Task Definition}

HLS prediction is formalized as a graph regression task. Following previous works~\cite{gnn-dse,harp}, we utilize the ProGraML graph to represent a source code~\cite{programl}, which is a control data flow graph. Nodes represent instructions, variables, and constant values, and edges represent the control flow and data flow. A GNN model is trained to predict the FPGA's latency and the utilization of four resources: LUT, FF, DSP, and BRAM. In the domain generalization setting, we train the model on $N$ source kernels $D^{(train)} = \{D_1, D_2, ..., D_N\}$, where $D_i = \{(X_i, T_{i1}, Y_{i1}), (X_i, T_{i2}, Y_{i2}), ..., (X_i, T_{i i_n}, Y_{i i_n})\}$ is the $i$-th kernel containing $i_n$ samples, and each sample consists of a program $X$, a pragma design $T$, and a label $Y$. For a new kernel $D_{test}$, we consider the constraint of data scarcity where we only use $k$ samples from the dataset $D^{(k)} \subset D_{test}, |D^{(k)}| = k$, to fine-tune the model. After fine-tuning, there are two ways of evaluation: offline evaluation and online evaluation. In the offline evaluation, we evaluate the MSE on the left-out test samples: $D_{test} \setminus D^{(k)}$. In the online evaluation, we do design space exploration (DSE) to search for good pragma designs. Based on the fine-tuned model's prediction, we select the top $M$ designs to run HLS, and evaluate the speedup ($M$ is set to 10 in our experiments).

\subsection{HARP Model}

The proposed hierarchical MoE can be adapted to any GNN model. We use one of the SOTA GNN models for HLS prediction, HARP~\cite{harp}, as the base model. Since the original ProGraML graph is not good at modeling long-range dependency, HARP creates a ``pseudo node'' for each basic block and connects it with every node within that basic block. A basic block is a sequence of instructions with a single entry point and a single exit point where the terminator instruction can be a branch, return, etc. We illustrate HARP's model structure in the appendix. We use $V$ to denote the set of all nodes and $V_B$ to denote the set of pseudo nodes. HARP consists of five components: (1) A GNN encoder that consists of 6 GNN layers and learns node representation; (2) pragma MLPs that use an MLP for each pragma type to transform the representation of pseudo nodes that are modified by that pragma type; (3) another GNN layer after pragma MLP that further updates the node representations; (4) graph pooling, which performs graph pooling on pseudo nodes $V_B$ to form one graph representation and on normal nodes $V \setminus V_B$ to form another, then concatenates the two; and (5) output MLP that makes the final prediction.

\section{Methods}

\begin{figure*}[tbp]
\centering
\includegraphics[width=\linewidth]{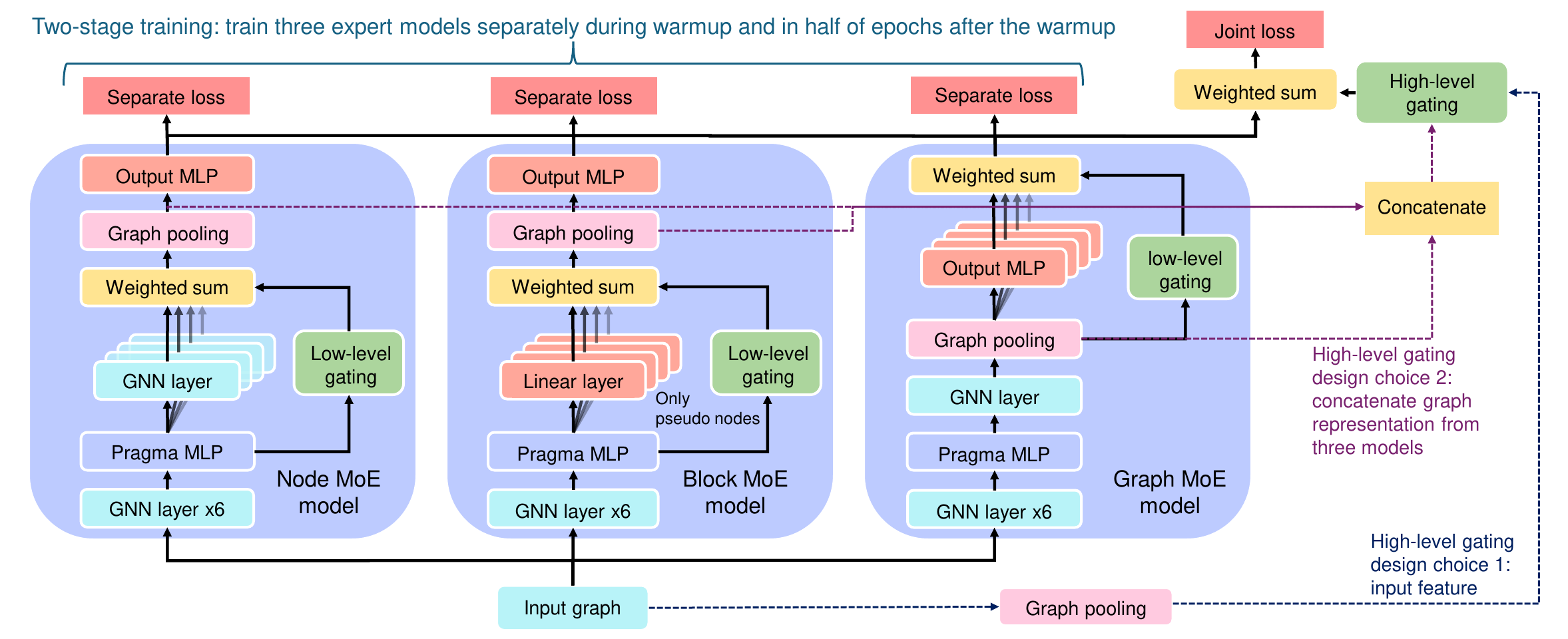}
\caption{Illustration of hierarchical MoE. The three low-level models are aggregated via the high-level gating network.}
\label{fig:model}
\vspace{-2mm}
\end{figure*}

\subsection{Low-Level Mixture of Experts}

A unique opportunity for the HLS prediction task is that the input graph has three granularities: normal nodes that represent data and instructions, pseudo nodes that represent basic blocks, and the whole graph that represents a source code file. When a data point from the target kernel shares a similar statement (node), basic block, or the whole code (graph) with a data point from the training kernels, then MoE is helpful. Thus, we consider employing MoE on the three granularities: MoE on all nodes including normal nodes and pseudo node (node MoE), MoE only on pseudo nodes (block MoE), and MoE on the graph (graph MoE). Based on our pre-exploration, the best practice is to apply MoE in the model components after the pragma MLP (the second component in HARP), since the experts need to share the same encoder and pragma MLPs. In our pre-exploration, we found that the best structure of the gating network is a linear layer. The model is shown in Figure~\ref{fig:model}.

\textbf{Node MoE.} For the MoE that operates on all nodes, we apply it on the GNN layer after the pragma MLP. We train $n$ GNN layers of the same structure as $n$ experts. We denote the representation of node $v_i$ after the pragma MLP as $\boldsymbol{h_i} \in \mathbb{R}^d$. The node MoE is formulated as:

\vspace{-2mm}
\begin{small}
\begin{equation}
\boldsymbol{h_i'} = \sum_{j=1}^n softmax(\boldsymbol{W_G^{node}} \boldsymbol{h_i})_j \cdot GNN_j(\boldsymbol{h_i}, \{h_j, j \in N(v_i)\}),
\label{equa:node_moe}
\end{equation}
\end{small}where $\boldsymbol{W_G^{node}} \in \mathbb{R}^{n \times d}$ is the parameter of the gating network, $N(v_i)$ is the set of neighbors of $v_i$, and $GNN_j(\cdot)$ is the $j$-th expert. The $softmax$ function is a normalization method that ensures that the weights of all the experts sum up to 1. The expert weights can be seen as ``attention.'' We calculate the expert weights (attention scores) using the gating network based on the node embeddings $\boldsymbol{h_i}$. By employing the MoE layer, different nodes can utilize different parameters for message passing and message aggregation.

\textbf{Block MoE.} We have tried applying MoE on the pragma MLP, but the performance decreased. It might be because each pragma type has a separate pragma MLP, so the training data for each pragma MLP is already limited. After we applied MoE, it diluted the training data even more. Instead, we add an additional MoE layer on the pseudo nodes' representations after the pragma MLP, so that we can increase the expressiveness of the pseudo nodes' representations before the graph pooling. The representation space of the pseudo nodes after the block MoE might be different from normal nodes, so it may not be suitable to have another GNN layer after the block MoE to do message-passing between pseudo nodes and normal nodes. We have also tried applying an additional layer for normal nodes after the pragma MLP, but the performance is unsatisfactory. It might be because the normal nodes lack a global view, so it is not useful to further transform its representation. After having the block MoE, doing graph pooling only on the pseudo nodes performs better than doing graph pooling on both pseudo nodes and normal nodes. Therefore, we do not utilize normal nodes after the GNN encoder in the block MoE model. Ablation study of various designs is in the appendix.

We use $n$ linear layers as $n$ expert networks. We denote the representation of pseudo node $v_i$ after the pragma MLP as $\boldsymbol{h_i} \in \mathbb{R}^d$. The block MoE is formulated as:

\vspace{-2mm}
\begin{equation}
\boldsymbol{h_i'} = \sum_{j=1}^n softmax(\boldsymbol{W_G^{block}} \boldsymbol{h_i})_j \cdot \boldsymbol{W_{j}}\boldsymbol{h_i} \ \  (v_i \in V_B),
\label{equa:block_moe}
\vspace{-1mm}
\end{equation} where $\boldsymbol{W_G^{block}} \in \mathbb{R}^{n \times d}$ is the parameter of the gating network, and $\boldsymbol{W_{j}} \in \mathbb{R}^{d \times d}$ is the $j$-th expert network.

\textbf{Graph MoE.} We apply the graph MoE on the output MLP. We denote the graph representation after graph pooling as $\boldsymbol{h_G}$. Different from the block MoE structure, here we use the original model components of the HARP model except for the output MLP. We employ $n$ output MLPs as $n$ expert networks. The graph MoE is formulated as:

\vspace{-2mm}
\begin{equation}
\hat Y^{(t)} = \sum_{j=1}^n softmax(\boldsymbol{W_G^{graph}} \boldsymbol{h_G})_j \cdot MLP^{(t)}_j(\boldsymbol{h_G}),
\label{equa:graph_moe}
\vspace{-1mm}
\end{equation} where $W_G^{graph} \in \mathbb{R}^{n \times d}$ is the parameter of the gating network, and $MLP^{(t)}_j(\cdot)$ is the $j$-th expert for predicting the objective $t$ ($t$ could be latency or a certain resource's utilization). The expert assignment is the same for all prediction objectives. The MLP contains four linear layers and the ELU activation function in between each. Different design points can utilize different experts to make the final prediction.

\textbf{Regularization term.} As discovered in previous research~\cite{shazeer2017,gmoe,moe_transfer_cv1}, expert polarization is a common problem of MoE. Due to the random initialization of expert networks, different experts initially perform differently. The gating network learns to assign higher weights to better experts. It results in more training of the initial better experts, leading to their even better performance, which is self-reinforcing. As training continues, the MoE model might collapse and only use the best expert. To avoid this problem, we apply a regularization term commonly used in MoE~\cite{shazeer2017}:

\vspace{-2mm}
\begin{small}
\begin{gather}
L_R = CV(\boldsymbol{I(W_G)}), \ \boldsymbol{I(W_G)} = \sum_{i=1}^M softmax(\boldsymbol{W_G} \boldsymbol{h_i}),
\label{equa:loss}
\end{gather}
\vspace{-2mm}
\end{small}

\noindent where $CV(\cdot)$ is the coefficient of variation. We calculate the importance score $\boldsymbol{I(W_G)} \in \mathbb{R}^n$ for gating network $W_G$ as the total weights assigned to each expert. For node/block MoE, $M$ is the total number of nodes/pseudo nodes in all graphs; for graph MoE, $M$ is the number of graphs. This regularization encourages balanced expert assignments.

\subsection{High-Level Mixture of Experts}

Now we have three MoE models operating on different granularities. A simple approach to combining them is to use all of them together in one model. However, our ablation study verifies that the performance will be worse. This demonstrates that we only need MoE on one granularity in one model. However, the best granularity greatly varies for different kernels, and it is difficult to discover a pattern. Therefore, we propose a high-level MoE to aggregate them. It calculates the weighted sum of the outputs of the three low-level MoE models.

We tested two design choices of the high-level gating network, as illustrated in Figure~\ref{fig:model}. The first design is to perform graph pooling on the input node features to form a graph representation as the input to the high-level gating network. We use the self-attention graph pooling. Denoting the input feature of node $v_i$ as $\boldsymbol{x_i}$, the graph pooling is formalized as:

\vspace{-2mm}
\begin{gather}
\boldsymbol{x_G} = \sum_{i \in V} softmax(MLP(\boldsymbol{x_i})) \cdot \boldsymbol{x_i},
\label{equa:input_gate}
\end{gather}
\vspace{-2mm}

\noindent where $X_G$ is the aggregated input feature, and $V$ is the set of all nodes.
The second design is to concatenate the graph representation in the three low-level MoE models.

If we use a sparse MoE where only one or two experts are selected, the first design would be more memory efficient. The first design can determine the expert assignments before the computation of three expert models, thus reducing unnecessary computation. However, the best-performing method is to utilize all the experts, and in this case, the two designs are similar in efficiency. The second design performs better in our experiments, since the hidden representations are more expressive than the input features, and so we utilize the second design in our experiments. We also apply the regularization term for the high-level gating network.

\subsection{Two-Stage Training Strategy}

Optimizing the hierarchical MoE is challenging. Different from previous MoE studies where expert networks have the same structures, our three low-level MoEs are very different and thus have different convergence speeds. The graph MoE model converges the fastest, since its MoE operates on the graph representation and has the least computation. Thus, the high-level gating network suffers more severely from expert polarization. It learns to assign nearly all the weights to the graph MoE model. If we simply increase the weight of the regularization term, the high-level gating network will learn to assign about $\frac{1}{3}$ weight to each expert, but the graph MoE model still converges the fastest. As a result, the graph MoE model will learn to output three times the prediction, while node and block MoE models will learn to output zero.

To address this unique challenge, we design a two-stage training strategy to encourage every expert model to perform well. In the first stage containing $T$ epochs (warmup), we train the three expert models individually. In the second stage, we take turns training the whole model end-to-end and the three expert models individually. If we denote the label as $Y$, the prediction made by the $i$-th expert model as $\hat{Y}_i$, and the MSE loss function as $L(Y, \hat{Y_i})$, then we define the loss function at epoch $t$ as:

\vspace{-2mm}
\begin{scriptsize}
\begin{equation}
L = \left\{
\begin{aligned}
&\frac{1}{3} \large[ L(Y, \hat{Y}_1) + L(Y, \hat{Y}_2) + L(Y, \hat{Y}_3) \large] + \alpha L_R, \ if \  t < T \ or \ 2 \mid t \\
&L(Y, \sum_{i=1}^3 g_i \cdot \hat{Y}_i) + \alpha L_R + \beta L_{Rh}, \ otherwise.
\end{aligned}
\right.
\label{equa:two_stage_loss}
\end{equation}
\end{scriptsize}
\vspace{-3mm}

Here, $L_R = \frac{1}{3} (L_{R1} + L_{R2} + L_{R3})$, where $L_{Ri}$ is the regularization term of the i-th low-level MoE. $L_{Rh}$ is the regularization term of the high-level MoE. $g_i$ is the weight assigned by the high-level gating network.

This two-stage training strategy does not increase the training time and is easy to implement. We only need to disable the high-level gating network and change the loss function in certain epochs. This strategy is only used during training. When we fine-tune a trained model on target kernels, we directly use the end-to-end joint training, since all the experts can already perform well and there is no risk of polarization. Besides, we initialize the high-level gating network to assign the same weights to the expert models, which can further prevent expert polarization. For low-level MoEs, we use the normal random initialization.


\begin{table*}[htbp]
    \centering
    \scalebox{0.96}{
    \begin{tabular}{cc|c|cccccccc}
    \toprule
    & & Offline evaluation & \multicolumn{8}{c}{Online evaluation (FPGA speedup compared to AutoDSE)} \\
    MoE category & Model & Total MSE & Fd & Gemv & Sy & Gemm & Ja & Tr & Average & Geo mean \\
    \midrule
    \multirow{3}{*}{No MoE} & HARP & 0.202$\pm$0.013 & 1.03 & 1.29 & 1 & 1 & \textbf{1.08} & 1.18 & 1.10 & 1.09 \\
    & HARP+MAML & 0.732$\pm$0.167 & 1 & 1 & 1 & 1 & 1 & 1 & 1 & 1 \\
    & ProgSG & 0.486$\pm$0.059 & 1 & 1 & 1 & 1 & 1 & 1.22 & 1.04 & 1.03 \\
    \midrule
    \multirow{3}{*}{Single MoE} & Node MoE & 0.160$\pm$0.035 & 1.13 & 1.03 & 1.00 & 1 & 1 & 1.15 & 1.05 & 1.05 \\
    & Block MoE & 0.171$\pm$0.019 & 1.43 & 1.11 & 1 & 1 & 1 & 1.07 & 1.10 & 1.09 \\
    & Graph MoE & 0.216$\pm$0.046 & 1 & 1.28 & \textbf{1.01} & 1 & \textbf{1.08} & \textbf{1.33} & 1.12 & 1.11 \\
    \midrule
    \multicolumn{2}{c|}{Hierarchical MoE} & \textbf{0.143$\pm$0.028} & \textbf{3.85} & \textbf{1.46} & 1 & \textbf{1.01} & 1 & 1.23 & \textbf{1.59} & \textbf{1.38} \\
    \bottomrule
    \end{tabular}
    }
    \caption{Domain generalization performance. ``Geo mean'' is the geometric mean speedup.}
    \label{table:main_results}
\vspace{-1.5mm}
\end{table*}

\section{Experiments}
\subsection{Experiment Settings}

\textbf{Datasets.} We use one of the most comprehensive benchmark datasets, HLSyn~\cite{hlsyn}. It contains 42 kernels covering various categories: linear algebra of vectors and matrices, data mining, stencil operations, etc. We utilize the AMD/Xilinx HLS tool, Vitis 2021.1~\cite{vitis}, to run HLS targeting the Xilinx Alveo U200 FPGA with a working frequency of 250MHz. We select 6 kernels as the target kernels that span representative categories including linear algebra, data mining, and stencil. The other kernels are source kernels. Table~\ref{table:dataset_statistics} shows the dataset statistics. We introduce their details in the appendix. The HLSyn dataset was generated by running the heuristics method, AutoDSE~\cite{AutoDSE}, for 24 hours. Many designs explored by AutoDSE are unavailable in HLS, so we collect this information to train a HARP classification model. Since its accuracy already exceeds 95\%, there is no need to employ MoE on the classification model. We use the available designs in the dataset to train the regression model, and we employ MoE on the regression model.

\begin{table}[tbp]
    \centering
    \scalebox{0.86}{
    \begin{tabular}{c c c c c c c c c}
    \toprule
    & Source & Fd & Gemv & Sy & Gemm & Ja & Tr \\
    \midrule
    \#Class & 37697 & 418 & 428 & 526 & 1421 & 1837 & 1651 \\
    \#Regre & 9418 & 77 & 231 & 121 & 348 & 257 & 103 \\ 
    \bottomrule
    \end{tabular}
    }
    \caption{Dataset statistics. ``\#Class'' and ``\#Regre'' denote the number of classification and regression data. The kernel acronyms represent ``fdtd-2d-large,'' ``gemver-medium,'' ``syr2k,'' ``gemm-p,'' ``jacobi-2d,'' and ``trmm-opt.''}
    \label{table:dataset_statistics}
\vspace{-2.5mm}
\end{table}

\textbf{Models.} Based on our pre-explorations, we use 4 experts in the low-level MoEs, and we set the regularization terms' weights of both low-level and high-level MoEs to 5e-3. Our baselines include (1) the SOTA GNN model, HARP~\cite{harp}; (2) HARP+MAML~\cite{gnn-dse-maml}, which applies the meta-learning method, MAML~\cite{MAML}, on HARP inspired by~\cite{gnn-dse-maml} to learn a more generalizable initialization of parameters; and (3) ProgSG~\cite{progsg}, which combines HARP and the language model. Hierarchical MoE brings many more parameters and needs about 32 GB of memory, but it can still fit into a single GPU. HARP and our model contain 359,370 and 1,329,403 parameters, respectively.

\textbf{Evaluation.} We train the regression and classification models on the source kernels. We want to mimic the domain transfer situation of having scarce but representative labeled data on the target kernels, so we use 50 data points per kernel to fine-tune our regression model, and roughly the same ratio of data points, 265 samples per kernel, to fine-tune the classification model. To select representative data points, we use K-means based on the graph representation. We conduct both offline and online evaluations. In offline evaluation, we calculate the fine-tuned regression model's mean squared error (MSE) on the left-out data points in the target kernels for each regression objective and sum up the five objectives' MSE.
In the online evaluation, we use the DFS search used in previous studies~\cite{gnn-dse,harp} to search for pragma designs. We limit our search range to 75,000 pragma designs, since it typically takes an hour. We use the fine-tuned classification model to predict the validity, and we use the fine-tuned regression model to predict the latency and resource utilization. From designs that are predicted to be valid and satisfy resource constraints, we choose the top 10 designs with the lowest predicted latencies to run HLS. We report the best design from the selected top-10 and the training dataset of the target kernels during fine-tuning, since these are all labeled data points, and we calculate its speedup compared to the best design in the dataset. We run the offline evaluation five times and the online evaluation three times, then we report the mean results.

\subsection{Experiment Results}

The experiments demonstrate MoE's effectiveness (shown in Table~\ref{table:main_results}). We use the second design of the high-level gating network as it performs better. MAML could not perform well in this setting. It might be because we have many more source kernels compared to the paper that proposes to use MAML on this task~\cite{gnn-dse-maml}. Different kernels might result in different directions of the meta gradient, leading to unstable training. ProgSG, which combines GNN and a language model, has a strong ability when the training data is sufficient, but it overfits in the data-scarce setting. During fine-tuning, the training loss is lower than 0.05, but the test loss is high. Comparatively, the MoE is more generalizable.

Hierarchical MoE is better than single MoE. In the online evaluation, on most kernels, the best of three single MoE models outperforms HARP, but different kernels favor different ones. For example, block is the best granularity for ``Fd',' while graph is the best granularity for ``Gemv,'' ``Ja,'' and ``Tr.'' Comparatively, hierarchical MoE performs the best or close to the best on almost every kernel.

\begin{table}[tbp]
    \centering
    \scalebox{0.85}{
    \begin{tabular}{c|ccc}
    \toprule
    & HARP & HARP+MAML & Hierarchical MoE \\
    \midrule
    MSE & 0.426 & 3.227 & \textbf{0.401} \\
    \bottomrule
    \end{tabular}
    }
    \caption{Offline evaluation results on more complex kernels.}
    \label{table:complex_kernels}
\vspace{-2.5mm}
\end{table}

\textbf{Complex kernels.} Since HLSyn kernels are generally simple, we test the model on 5 more complex kernels and verified the effectiveness of Hierarchical MoE. The 5 kernels include "3d-rendering" and "spam-filter" from the Rosetta benchmark~\cite{rosetta} and three self-constructed kernels that are introduced in the appendix. These are closer to real-world applications. Following previous settings, we train the regression model on HLSyn kernels and finetune it on 50 points per kernel on the new kernels. Table~\ref{table:complex_kernels} shows the results. It is challenging for all models to perform well, but the hierarchical MoE model performs the best. It remains a future work to improve models on complex kernels.

\subsection{Ablation Study}

\begin{table}[tbp]
    \centering
    \scalebox{0.94}{
    \begin{tabular}{c|ccccc}
    \toprule
    Data split & N+B & N+G & B+G & N+B+G & Hierarchy \\
    \midrule
    K-means & 0.341 & 0.213 & 0.188 & 0.174 & \textbf{0.143} \\
    Random & 0.893 & 0.562 & 3.126 & 0.984 & \textbf{0.452} \\
    \bottomrule
    \end{tabular}
    }
    \caption{Using MoE on various granularities in a single model. ``N,'' ``B,'' and ``G'' represent node, block, and graph.}
    \label{table:ablation_study_single}
\end{table}

To verify the effectiveness of our model design, we do ablation studies on different model structures. First, instead of aggregating the three low-level MoEs, we try applying MoE on the three granularities together in a single model, and we specify the detailed structure design in the appendix. Table~\ref{table:ablation_study_single} shows the results. When we use MoE on two or three granularities in a single model, the loss is not only higher than hierarchical MoE but even higher than the lowest loss when only using MoE on one granularity. Apart from selecting 50 representative data points by K-means for fine-tuning, we also experiment with random data split. Hierarchical MoE is the most stable model when it faces low-quality fine-tuning data in the random split. If we stack MoE on various granularities in the same model, the structure might be too complex and thus unstable to train on a small dataset. The results demonstrate the effectiveness of hierarchical MoE.

\begin{table}[tbp]
    \centering
    \scalebox{0.88}{
    \begin{tabular}{c|cccc}
    \toprule
    Metric & Node & Block & Graph & All \\
    \midrule
    Offline (MSE) & 0.164 & 0.177 & 0.279 & \textbf{0.143} \\
    Online (geo mean speedup) & 1.34 & 1.13 & 1.35 & \textbf{1.38} \\
    \bottomrule
    \end{tabular}
    }
    \caption{Using MoE on a single granularity in the hierarchical MoE. ``Node/Block/Graph'' means only using MoE on the node/block/graph granularity in the hierarchical MoE.}
    \label{table:ablation_study_hier}
\vspace{-2.5mm}
\end{table}

Second, we want to examine whether the hierarchical MoE benefits from combining the three granularities. To answer it, we still use the hierarchical MoE structure, but we only use one low-level MoE's granularity. Table~\ref{table:ablation_study_hier} shows the results. It verifies that aggregating the three granularities performs the best. Using only the node MoE or graph MoE also performs well in the online evaluation, and this might be due to the increased number of experts. Nonetheless, aggregating the three granularities further improves the performance.

\begin{table}[tbp]
\vspace{-1.5mm}
    \centering
    \scalebox{0.79}{
    \begin{tabular}{c|cc|cc|cc}
    \toprule
     & \multicolumn{2}{c|}{Two-stage training} & \multicolumn{2}{c|}{W/o alternative train} & \multicolumn{2}{c}{W/o warmup} \\
    Gating & Input & Hidden & Input & Hidden & Input & Hidden \\
    \midrule
    MSE & 0.149 & \textbf{0.143} & 0.193 & 0.159 & 0.152 & 0.184 \\
    \bottomrule
    \end{tabular}
    }
    \caption{Ablation study of two-stage training and high-level gating network. ``Input'' is the first design of the high-level gating network, while ``Hidden'' is the second design.}
    \label{table:ablation_study}
\vspace{-1.5mm}
\end{table}

Third, we want to verify whether the two-stage training and the high-level gating network's second design are effective. The results are reported in Table \ref{table:ablation_study}. The two-stage training contains two stages: (1) the warmup stage where we train the three experts separately, and (2) training the three experts jointly and separately in turn. If we disable either one of them, the loss will increase. Also, the high-level gating network's second design (based on hidden representations) is better than its first design (based on input features).


More ablation studies are in the appendix. They verify that the hierarchical MoE's performance is not due to increased parameter size or expert number, but due to the hierarchical structure and combination of three granularities.

\subsection{Analysis of Expert Assignment}

\begin{table}[tbp]
    \centering
    \scalebox{0.89}{
    \begin{tabular}{c|cccccc}
    \toprule
    Expert & Fd & Gemv & Sy & Gemm & Ja & Tr \\
    \midrule
    Node MoE & 37\% & 29\% & 36\% & 37\% & \textbf{46\%} & 27\% \\
    Block MoE & \textbf{49\%} & 28\% & 26\% & \textbf{40\%} & 32\% & \textbf{37\%} \\
    Graph MoE & 14\% & \textbf{43\%} & \textbf{38\%} & 23\% & 22\% & 36\% \\
    \bottomrule
    \end{tabular}
    }
    \caption{Average assigned weights of the high-level MoE.}
    \label{table:expert_assignment}
\end{table}

We want to unveil the mystery of the gating networks. We find that their output weights are partially explainable to the model's performance. We show the average assigned weights by the high-level gating network in Table~\ref{table:expert_assignment}. According to Table~\ref{table:main_results}, ``Fd,'' ``Gemv,'' and ``Tr'' have a strong preference for a certain granularity, while the other three kernels do not. Among them, the block MoE performs the best for ``Fd'' and it is also assigned the highest weight; graph MoE is the best expert for ``Gemv'' and ``Tr,'' and it is also assigned the highest or nearly the highest weight.

\begin{table}[tbp]
    \centering
    \scalebox{0.73}{
    \begin{tabular}{c|ccccc}
    \toprule
    Expert & No pragma & Loop tiling & Pipeline & 0$\textless$Parallel$\leq$4 & Parallel$\textgreater$4 \\
    \midrule
    1 & \textbf{32\%} & 6\% & 6\% & 4\% & 2\% \\
    2 & 18\% & 8\% & 13\% & 41\% & \textbf{81\%} \\
    3 & 22\% & \textbf{64\%} & \textbf{60\%} & \textbf{43\%} & 8\% \\
    4 & 28\% & 22\% & 21\% & 12\% & 10\% \\
    \bottomrule
    \end{tabular}
    }
    \caption{Average assigned weights of the block MoE. Each column shows the expert weights for pseudo nodes modified by a certain pragma type.}
    \label{table:analysis_block}
\vspace{-2.5mm}
\end{table}

Besides, we find that low-level gating networks are also explainable. We pick the best hierarchical MoE model based on the previous five repeated offline evaluation experiments to do a case study on the low-level MoE. Due to the limited space, here we only analyze the block MoE, and we analyze the node and graph MoEs in the appendix. The pragmas modify the pseudo nodes, so block MoE is a window for us to analyze the pragmas. We summarize the weights of each expert for each pragma type in Table~\ref{table:analysis_block}. There are three types of pragmas: loop tiling, pipeline, and parallelization. The third expert is good at dealing with loop tiling, pipeline, and small parallel factors; the second expert is good at dealing with large parallel factors; the other two experts deal more with pseudo nodes that do not have pragmas. Different experts diversify their roles, which improves generalizability.

\section{Conclusion}

Domain generalization is a big challenge for HLS prediction models. We propose the hierarchical MoE structure. In the low-level MoE, we apply MoE on one of the three natural granularities of the graph: node, basic block, or graph. In the high-level MoE, we aggregate the three low-level MoE models, so that different data points can flexibly use one of them. To address the expert polarization, we propose a two-stage training strategy. Extensive experiments have verified our model's effectiveness. Meanwhile, the generalizability of HLS prediction models still remains a big challenge, and future works should further improve the performance.

\section*{Acknowledgments}
This work was partially supported by NSF grants 2211557, 1937599, 2119643, 2303037, and 2312501, SRC JUMP 2.0 PRISM Center, Amazon Research, Snapchat, and the CDSC industrial partners (https://cdsc.ucla.edu/partners/). The authors would also like to thank AMD/Xilinx for HACC equipment donation, and Marci Baun for editing the paper. J. Cong has a financial interest in AMD.

\bibliographystyle{aaai}
\bibliography{reference}

\clearpage
\newpage
\appendix
\section{Additional Background Knowledge of HLS}

\begin{figure}[htbp]
\includegraphics[width=\linewidth]{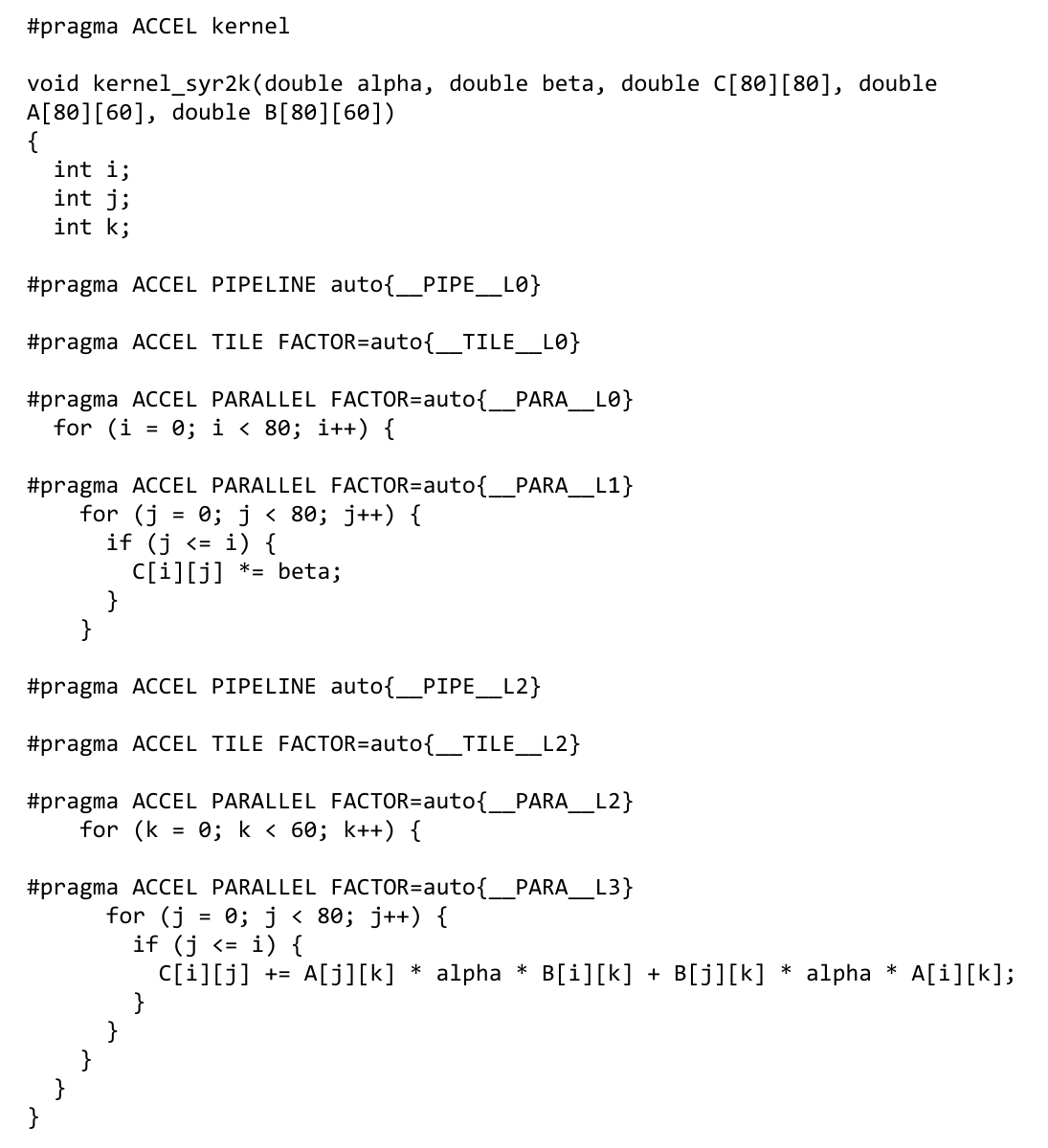}
\caption{Source code of the ``syr2k'' kernel. The source code consists of the program that describes the FPGA's function, and several pragmas inserted in the program that instruct the hardware synthesis process, such as parallelization, pipeline, etc. ``auto\{pragma index\}'' are pragma's placeholders. We utilize the machine learning model to find the best pragmas to insert in these placeholders.}
\label{fig:syr2k_example}
\end{figure}

As the HLS prediction task is quite new to the AI community, we would like to illustrate this task at the beginning of the appendix. Taking the ``syr2k'' kernel as an example, Figure~\ref{fig:syr2k_example} shows its source code. As explained in the Introduction section, the source code consists of the program that describes the FPGA's function, which is similar to any normal C program, and several pragmas. Now the pragmas are represented by the placeholders ``auto\{pragma index\}.'' We utilize the machine learning model to find the best pragmas to insert in these placeholders.

Following HARP~\cite{harp}, we develop the model based on the open-source AMD/Xilinx Merlin Compiler~\cite{merlin}, because it reduces the complex HLS pragmas to only three Merlin pragmas, greatly benefiting the pragma search process. It applies source-level code transformations to enable various automatic architectural optimizations such as memory burst, memory coalescing, and coarse-grained optimizations. There are three types of Merlin pragmas: parallel, pipeline, and loop tiling. For the parallel pragma placeholders, we should insert an integer as the parallel factor. For the pipeline pragma placeholders, we should choose between the fine-grained pipeline, coarse-grained pipeline, or not using the pipeline. For the loop tiling pragma placeholders, we should insert an integer as the loop tiling factor.

Designing the pragmas is labor-intensive and heavily relies on hardware knowledge, which motivates the automated pragma design research. Different pragmas can lead to significantly different FPGA performance. Table~\ref{table:dataset_latency} shows the latency (clock cycle) of each target kernel's best pragma design and worst pragma design in the HLSyn dataset. The shorter latency the better. We also list the mean and standard deviation of latency in the dataset, and the latency of not inserting any pragma. We can see that the gap between the best design and worst design is very big, and the worst design is sometimes even worse than not inserting any pragma. This is because some pragmas might use too many resources and occupy the resources of other modules. The standard deviation is also very large, and sometimes it is even larger than the mean. Therefore, choosing a good pragma design is very important to the FPGA's performance.

\begin{table*}[tbp]
    \centering
    \scalebox{0.81}{
    \begin{tabular}{c|cccccc}
    \toprule
    & Fdtd-2d-large & Gemver-medium & Syr2k & Gemm-p & Jacobi-2d & Trmm-opt \\
    \midrule
    Best design & 2,355,778 & 210,335 & 46,061 & 9,179 & 164,284 & 9,387 \\
    Worst design & 96,496,392 & 1,826,012 & 2,005,329 & 2,110,866 & 5,520,201 & 1,547,247 \\
    Mean & 27,529,638+-37,088,252 & 644,066+-313,804 & 170,139+-353,609 & 135,877+-251,531 & 3,049,134+-1,544,708 & 329,745+-347,427 \\
    No pragma & 14,362,849 & 1,826,012 & 533,981 & 461,231 & 625,124 & 1,547,247 \\
    \bottomrule
    \end{tabular}
    }
    \caption{The latency (clock cycle) of each target kernel's best pragma design and worst pragma design in the HLSyn dataset. We also list the mean and standard deviation of latency in the HLSyn dataset, and the latency of not inserting any pragma.}
    \label{table:dataset_latency}
\end{table*}

\section{Additional Problem Analysis}

\subsection{Challenges of Domain Shift}
\begin{figure}[htbp]
\centering
\includegraphics[width=0.8\linewidth]{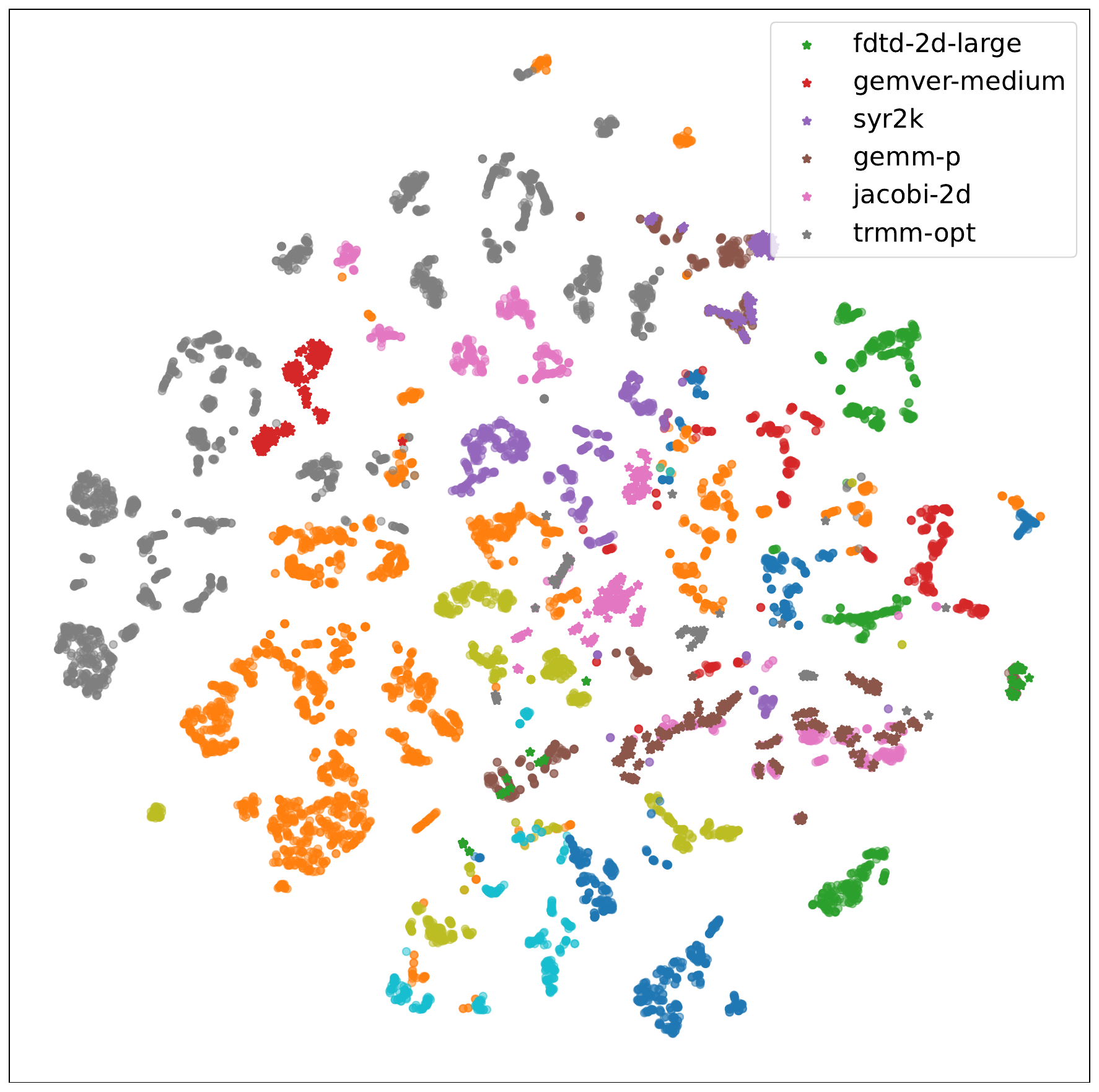}
\caption{Visualization of the graph embeddings. Each point is a pragma design. Points belonging to the same kernel have the same color. Points from the same kernel are clustered together, while points from different kernels are located in different places. Due to the limited number of colors, different kernels might use the same color. Due to the limited space, we only show the names of the target kernels.}
\label{fig:intro_visualization}
\end{figure}

Domain shift brings a lot of challenges to our task. First, different kernels have large differences. As shown in Table~\ref{table:dataset_latency}, the mean latency of different kernels is very different. We hope the model can learn this distribution, so we cannot align the representation space of different kernels, as did in many previous domain generalization papers. We extract the graph representation of each design point from the HARP model and visualize it in Figure~\ref{fig:intro_visualization}. In the figure, points belonging to the same kernel have the same color. Due to the limited number of colors, different kernels might use the same color. We can see that points from the same kernels are clustered together, while points from different kernels are located in different places. The visualization further verifies that the domain difference between different kernels is very big. Such a big difference makes the domain transfer very difficult.

\begin{figure}[htbp]
\includegraphics[width=0.75\linewidth]{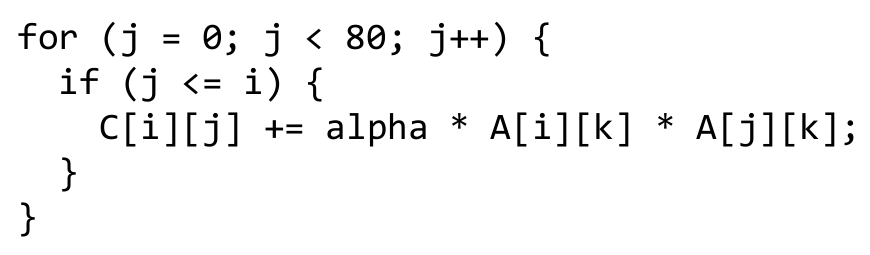}
\caption{Code snippet of the ``syrk'' kernel. The line inside this nested loop is the only difference between the ``syr2k'' kernel and the ``syrk'' kernel. However, their best pragma designs are very different.}
\label{fig:syrk_example}
\end{figure}

Second, labeling the new kernel's data points by HLS is time-consuming. Each run takes minutes to hours~\cite{harp}. Therefore, we use the constraint of data scarcity of only using 50 data points per kernel to finetune the model on new kernels.

Nonetheless, a unique opportunity is that some different kernels share some similar substructions, where both their similarities and differences should be carefully noticed. The similarity could benefit the MoE model by utilizing the knowledge learned from other kernels. Nonetheless, the difference is also very important. An example is the ``syr2k'' kernel and the ``syrk'' kernel. They only differ in one line inside a nested loop, shown in Figure~\ref{fig:syrk_example} and Figure~\ref{fig:syr2k_example}. However, their best pragma designs are very different. In the best pragma design of ``syrk,'' the tiling factor of ``\_\_TILE\_\_L2'' is 60, but it is 1 for ``syr2k''; the parallel factor of ``\_\_PARA\_\_L1'' and ``\_\_PARA\_\_L2'' are 20 and 1 for ``syrk,'' but they are 8 and 4 for ``syr2k.'' It is challenging for the model to not only benefit from the similarities and also observe the slight differences.
Therefore, domain generalization is difficult for HLS prediction models. The MoE structure can utilize similar experts to deal with the similarities, but it can also discern the differences and use different experts to deal with them.

\subsection{HARP Model Illustration}

\begin{figure}[htbp]
\centering
\includegraphics[width=0.3\linewidth]{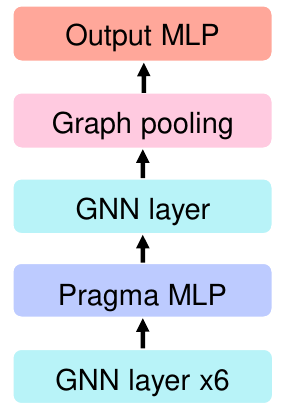}
\caption{HARP model diagram.}
\label{fig:harp}
\end{figure}

Figure~\ref{fig:harp} illustrates the HARP model using the same diagram style as our hierarchical MoE model's diagram (Figre~\ref{fig:model}). It shows the five components of the HARP model. This diagram ignores some details of the HARP model. For more details, please refer to the HARP paper~\cite{harp}.

\section{Detailed Experiment Settings}

\textbf{Datasets.} The HLSyn benchmark dataset~\cite{hlsyn} consists of 42 kernels selected from the MachSuite dataset~\cite{machsuite} and the Polyhedral dataset (PolyBench)~\cite{polybench}. Since the FPGA's latency is too diverse, it is difficult for the model to directly learn to predict. Thus, we normalize the FPGA's latency by $0.5  * log_2 (\frac{1e7}{\text{latency}})$ as the normalized performance. It is the lower the better for the original latency, but it is the higher the better for the normalized performance. Our HLS targets the Xilinx Alveo U200 FPGA with a working frequency of 250MHz. We also normalize the resource utilization by dividing it by the total available resources of the targeting FPGA. We mainly care about four types of resources: LUT, FF, DSP, and BRAM, so we have five prediction targets including the normalized performance.

Our resource constraint is that each resource type should be used no more than 80\%, because we need to leave some spaces for downstream tasks or other additional modules. We hope the model can make accurate predictions of the performance and resource utilization, so we can search for the best pragma design that satisfies the resource constraints and has the highest normalized performance (lowest latency). We rely on the classification model to help us wipe out the invalid designs, but we rely on the regression model to help us wipe out the valid designs that exceed our resource constraints.

\textbf{More Complex Kernels.} Since the kernels in the HLSyn benchmark dataset~\cite{hlsyn} are relatively small and simple, we also conduct experiments on more complex kernels that are closer to real-world applications. We select two kernels from the Rosetta benchmark dataset~\cite{rosetta}: 3d-rendering and spam-filter. The ``3d-rendering'' kernel renders 2D images from 3D triangle mesh models. The ``spam-filter'' kernel trains a logistic regression model for spam email classification. The details of these two kernels can be found in the Rosetta benchmark paper.

We also construct three new kernels by ourselves. The first kernel, ``att-3mm,'' calculates the attention mechanism widely used in the Transformer~\cite{Transformer} architecture. Its input is an embedding matrix and two parameter matrices. It uses the embedding matrix and the first parameter matrix to calculate the Q matrix, uses the embedding matrix and the second parameter matrix to calculate the K matrix, and then multiplies Q and K to get the attention matrix. The second kernel, ``att-3mm-fuse,'' does the same calculation but it fuses the outer loop. The third kernel, ``vmmv,'' is constructed by combining vector-matrix multiplication and matrix-matrix multiplication inspired by the ``atax'' kernel and the ``gemm-p'' kernel in the HLSyn benchmark.

Following the previous settings, we train the regression model on the complete HLSyn data and finetune it on 50 points per kernel on the new kernels. We use K-means to select 50 data points per kernel. We list these complex kernels' statistics in Table~\ref{table:complex_kernel_statistics}.

\begin{table}[h]
    \centering
    \scalebox{0.93}{
    \begin{tabular}{c c c c c c c c}
    \toprule
    & HLSyn & 3d & Spam & Att & Att-fuse & Vmmv \\
    \midrule
    \#Regre & 10555 & 124 & 520 & 90 & 139 & 298 \\
    \bottomrule
    \end{tabular}
    }
    \caption{Dataset statistics of the complex kernels. ``\#Regre'' denotes the number of regression data. The kernel acronyms represent ``3d-rendering,'' ``spam-filter,'' ``att-3mm,'' ``att-3mm-fuse,'' and ``vmmv.''}
    \label{table:complex_kernel_statistics}
\end{table}

\textbf{Evaluation.} During pre-training on the source kernels, we use 80\% of data points as the training set, 10\% for validation, and 10\% for testing. We use five random seeds (1, 2, 3, 4, 5) to randomly split the source kernel's dataset. The MSE on the source kernels is unimportant. We pretrain the model for 1000 epochs, and we select the model checkpoint with the lowest validation MSE on the validation set. Then, we finetune it for 500 epochs on the target kernels. We use 50 data points per kernel as the training set during fine-tune, and the other data points are the test set. We report the test MSE during the epoch which has the lowest training MSE. During fine-tuning, we use the same random seed as used in that model's pretraining. We run offline evaluation five times. For online evaluation, since HLS takes a long time, we only use the three models trained with random seeds 1, 2, and 3 to run HLS. In offline evaluation, we calculate the sum of the MSE of the five prediction targets on the target kernels' test set. In online evaluation, from the designs that are predicted to be valid and under resource constraints, we pick the top 10 designs with the lowest predicted latencies (highest predicted normalized latencies) to run HLS. When running HLS, we set the timeout as 3 hours. If the HLS does not finish within 3 hours, it is very likely to be unsynthesizable, so we will terminate the HLS and treat it as an invalid design. We report the best design from the selected top-10 and the training dataset of the target kernels, since these are all labeled data points, and we calculate its speedup compared to the best design in the dataset. We calculate the mean speedup of the three repeated HLS experiments for each target kernel. Then we calculate the geometric mean speedup for all target kernels.

\textbf{MoE on various granularities in a single model.} In the experiments section in the main paper, we conduct an experiment of applying MoE on various granularities in a single model (Table~\ref{table:ablation_study_single} in the main paper). When we utilize the block MoE and the graph MoE together, or the node MoE and the graph MoE together, there is no conflict and we can smoothly apply them. However, when we utilize node MoE and block MoE together, or when we utilize MoE on all three granularities, we face a conflict. The block MoE model removes the GNN layer after the pragma MLP, but the node MoE is applied to that GNN layer. Therefore, we slightly modify the model structure to accommodate both models. We keep that GNN layer after the pragma MLP and apply MoE on it, as the node MoE. After it, we follow the paradigm of the block MoE model: adding an additional block MoE layer for pseudo nodes after that GNN layer, and we do graph pooling only on pseudo nodes. We can further apply graph MoE on the output MLP. In this way, we allow MoE on both node and block granularity, or on all three granularities.

\textbf{Baseline methods.} We have three baseline models: HARP~\cite{harp}, HARP+MAML~\cite{MAML}, and ProgSG~\cite{progsg}. For our models, we follow the same hyper-parameter setting and training strategy as HARP, except for the MoE's unique hyper-parameters. We will specify the hyper-parameters in a later section.

\begin{itemize}
\item HARP~\cite{harp}: we use its original settings and official codes (https://github.com/UCLA-VAST/HARP).

\item HARP+MAML: inspired from the paper~\cite{gnn-dse-maml}, we use the meta-learning method, MAML~\cite{MAML}, to help find a good initialization for the HARP model's parameters based on the source kernels. Then we use the normal fine-tuning strategy on the target kernels. We use the implementation of the ``learn2learn'' Python package. We use the same hyper-parameter setting as HARP. For the MAML part, we tune its hyper-parameters and select the hyper-parameters as follow: we randomly sample 32 kernels with replacement per epoch to train, and we use 80\% of data as the training set.

\item ProgSG~\cite{progsg}: it uses both text embedding of the source code and the ProGraML graph as the input. We use its official codes (https://github.com/zongyueqin/progsg).
\end{itemize}

\section{Supplementary Experiment Results}

\subsection{Study of Model Variants}

One concern is whether the hierarchical MoE model simply benefits from having more parameters. To address this concern, we enlarge the hidden size of HARP from 64 to 128 and 256, and we evaluate their performance. The results are listed in Table~\ref{table:ablation_study_size}. As we enlarge the hidden size, the MSE increases, so its performance becomes worse. In the original best hyper-parameter setting of HARP, 64 is already the best-performing hidden size. We also list the parameter size of single-level MoE models in Table~\ref{table:model_size}. If we do not utilize the hierarchical structure, only using MoE on one granularity does not significantly increase the parameter size, but it decreases the MSE. It verifies that parameter size is not the key reason for MoE's success.

\begin{table}[h]
    \centering
    \scalebox{0.9}{
    \begin{tabular}{cc|ccc}
    \toprule
    Model & Hidden size & \#Param & MSE \\
    \midrule
    \multirow{3}{*}{HARP} & 64 & 359,370 & 0.202 \\
    & 128 & 1,053,610 & 0.240 \\
    & 256 & 3,445,130 & 0.327 \\
    \midrule
    Hierarchical MoE & 64 & 1,329,403 & \textbf{0.143} \\
    \bottomrule
    \end{tabular}
    }
    \caption{Influence of the model's parameter size. ``\#Param'' indicates the parameter size.}
    \label{table:ablation_study_size}
\end{table}

\begin{table}[h]
    \centering
    \scalebox{0.9}{
    \begin{tabular}{c|cccc}
    \toprule
     & Node MoE & Block MoE & Graph MoE \\
    \midrule
    \#Param & 473,870 & 361,805 & 431,901 \\
    \bottomrule
    \end{tabular}
    }
    \caption{Parameter size of the single-level MoE models that do not have the hierarchical structure.}
    \label{table:model_size}
\end{table}

Another concern is about the number of experts. In the hierarchical MoE, if we only utilize MoE on the node granularity or graph granularity, its performance is slightly worse than utilizing MoE on three granularities, but it could still perform well (Table~\ref{table:ablation_study_hier} in the main paper). In this case, the three expert models of the high-level MoE have the same structure. Each of the low-level MoE models has 4 experts, so in total we have 12 experts for that granularity and 3 experts for other layers that are not equipped with low-level MoE. Does its success come from a large number of expert models (12 experts for that granularity), or the hierarchical structure (12 experts for that granularity + 3 experts for other layers)? To answer this question, we run experiments on the single-level MoE structure with different numbers of experts. In addition to our original setting using 4 experts, we experiment with 2, 8, and 12 experts. The results are listed in Table~\ref{table:study_num_experts}. The results demonstrate that increasing the number of experts does not necessarily improve the performance. For all three MoEs, the lowest MSE is achieved when we use 4 experts. Therefore, the performance gain is probably because of the hierarchical structure (12 experts for that granularity and 3 experts for other layers).

\begin{table}[h]
    \centering
    \scalebox{0.91}{
    \begin{tabular}{c|cccc}
    \toprule
    Model & 2 experts & 4 experts & 8 experts & 12 experts \\
    \midrule
    Node MoE & \textbf{0.160} & \textbf{0.160} & 0.302 & 0.195 \\
    Block MoE & 0.208 & \textbf{0.171} & 0.207 & 0.264 \\
    Graph MoE & 0.305 & \textbf{0.216} & 0.229 & 0.229 \\
    \bottomrule
    \end{tabular}
    }
    \caption{Influence of the number of experts in the single-level MoE model. The listed results are test MSE.}
    \label{table:study_num_experts}
\end{table}

As discussed in the main paper, when we were designing the block MoE's structure, there were many design choices, from which we chose the current one. We also conduct experiments on other variants of the single-level block MoE model. In our design, we remove the GNN layer after the pragma MLP, and we remove the graph pooling on the normal nodes. Therefore, we only rely on the pseudo nodes after the GNN encoder. This is because the pseudo nodes are sufficient for making the final prediction after we add the block MoE layer. Here we preserve the GNN layer after the pragma MLP, or preserve both the GNN layer and the graph pooling on normal nodes. The results are shown in Table~\ref{table:ablation_block_moe}. Our design archives the lowest MSE. It verifies the effectiveness of our design.

\begin{table}[h]
    \centering
    \scalebox{0.79}{
    \begin{tabular}{c|ccc}
    \toprule
    & Block MoE & Block MoE+GNN & Block MoE+GNN+pooling \\
    \midrule
    MSE & \textbf{0.171} & 0.529 & 0.240 \\
    \bottomrule
    \end{tabular}
    }
    \caption{Variants of the single-level block MoE model. ``Block MoE+GNN'' means we preserve the GNN layer after the pragma MLP in the block MoE model. ``Block MoE+GNN+pooling'' means we preserve the GNN layer after the pragma MLP and the graph pooling on normal nodes in the block MoE model.}
    \label{table:ablation_block_moe}
\end{table}

\subsection{Analysis of Expert Assignment}

It is interesting to see whether the expert assignment is meaningful. Due to the limited space, we only analyze the gating network of the high-level MoE and the block MoE in the main paper. Here we analyze the gating network of the node MoE and the graph MoE. We still pick the best-performing hierarchical MoE model from the five repeated experiments to analyze. For the node MoE, we show the averaged assigned weights for each type of node in Table~\ref{table:analysis_node}. There are three non-overlapping types of nodes: normal nodes that represent the data and instructions, pseudo nodes that represent basic blocks, and pragma nodes that represent pragmas. A special type of normal node is icmp node, which represents the ``icmp'' instruction of a loop. If a pragma modifies a loop, that pragma node will be connected to both the icmp node and the pseudo node corresponding to the loop, so the icmp nodes are very meaningful. The normal nodes and pseudo nodes are relatively equally distributed to the experts. The icmp nodes are mainly handled by the third expert, and the pragma nodes are mainly handled by the fourth expert. Similar to our analysis of the block MoE in the main paper, the node MoE's experts also diversify their roles.

\begin{table}[h]
    \centering
    \scalebox{0.9}{
    \begin{tabular}{c|cccc}
    \toprule
    Expert & Normal & Icmp ($\in$ normal) & Pragma & Pseudo \\
    \midrule
    1 & 25\% & 0\% & 25\% & 30\% \\
    2 & 23\% & 31\% & 0\% & 27\% \\
    3 & \textbf{27\%} & \textbf{69\%} & 10\% & 11\% \\
    4 & 25\% & 0\% & \textbf{65\%} & \textbf{32\%} \\
    \bottomrule
    \end{tabular}
    }
    \caption{Average assigned weights of the node MoE. Each column shows the expert weights for a certain type of node. There are three non-overlapping types of nodes: normal nodes, pseudo nodes (basic block), and pragma nodes.}
    \label{table:analysis_node}
\end{table}


\begin{table*}[h]
    \centering
    \scalebox{0.95}{
    \begin{tabular}{cc|cc}
    \toprule
    Target kernel & Domain & Source kernel & Source domain \\
    \midrule
    fdtd-2d-large & 2-D finite different time domain kernel & bicg-large & BiCG sub kernel of BiCGStab linear solver \\
    gemver-medium & matrix-vector multiplication & gesummv-medium & matrix-vector multiplication \\
    syr2k & symmetric rank-2k operations & syrk & symmetric rank-k operations \\
    gemm-p & matrix multiplication & covariance & covariance computation \\
    jacobi-2d & 2-D Jacobi stencil operations & doitgen-red & multiresolution analysis \\
    trmm-opt & triangular matrix multiplication & symm & symmetric matrix multiplication \\
    \bottomrule
    \end{tabular}
    }
    \caption{The most similar source kernel of each target kernel based on the graph MoE's expert assignment weights.}
    \label{table:analysis_graph}
\end{table*}

\begin{table*}[h]
    \centering
    \scalebox{0.91}{
    \begin{tabular}{c|c|c}
    \toprule
    Hyper-parameter & Scope & Choice \\
    \midrule
    Number of experts & 2, 4, 8, 12 & 4 \\
    Number of used experts & 2, 4, 8, 12 & 4 \\
    Regularization term $\alpha$ & 5e-4, 1e-3, 5e-3, 1e-2 & 5e-3 \\
    Regularization term $\beta$ & 5e-4, 1e-3, 5e-3, 1e-2 & 5e-3 \\
    Fine-tuning epochs & 300, 500, 700 & 500 \\
    Node MoE's place & every GNN layer in the GNN encoder, the GNN layer after pragma MLP*, additional linear layer & * \\
    Block MoE's place & pragma MLP, pooling, additional linear layer, additional linear layer and remove some layer* & * \\
    \bottomrule
    \end{tabular}
    }
    \caption{The scope of MoE-related hyper-parameters that we have tried. ``Choice'' is the best choice (our final choice).}
    \label{table:hyper_parameter}
\end{table*}

For graph MoE, based on the expert assignment weights of each data point, we calculate the average expert assignment weights of each kernel. Then, we can calculate the cosine similarity between two kernels using the weights. Table~\ref{table:analysis_graph} lists the most similar source kernel for each target kernel. ``Gemver-medium'' and its most similar kernel, ``gesummv-medium,'' are both matrix-vector multiplications; ``syr2k'' and its most similar kernel, ``syrk,'' are both symmetric rank operations; ``trmm-opt'' and its most similar kernel, ``symm,'' are both matrix multiplications. This indicates that the expert weights assigned by the graph MoE are partially explainable.

\section{Other Detailed Information}

\textbf{Hyper-parameters.} We basically follow the hyper-parameter settings of HARP. We use the Adam optimizer to train the model for 1000 epochs and fine-tune it for 500 epochs. The learning rate is 5e-4 and the weight decay is 1e-4. We use the cosine annealing learning rate schedule~\cite{SGDR_lr_scheduler} with a minimum learning rate of 1e-5, and the unturned linear warmup~\cite{untuned_warmup}. We set the hidden size as 64 and the dropout as 0.1. The regression models use the MSE loss to train, while the classification models use the cross entropy loss. In the first design of the high-level gating network, we use the graph pooling of input features (Equation~\ref{equa:input_gate} in the main paper). Its MLP has two layers and one ReLU function between them, where the first layer has the same hidden size as the input dimension and the second one shrinks the output dimension to 1.

For hyper-parameters related to MoE, we try the hyper-parameters in a reasonable scope and use the best one with the lowest training-set MSE on the target kernels. The low-level MoE uses the weighted sum of 4 experts, and the high-level MoE calculates the weighted sum of three models (the node MoE model, the block MoE model, and the graph MoE model). The gating network is a linear layer and a softmax function. We set the weight of the regularization term of MoE, both $\alpha$ and $\beta$, as 5e-3. In the two-stage training of the high-level MoE, the first stage (warmup) takes the first 500 epochs, and the second stage takes the later 500 epochs. In the model variants' experiments, we only modify the changing hyper-parameters as introduced in the experiments. Table~\ref{table:hyper_parameter} lists the scope of hyper-parameters we have tried.

\textbf{Reproducibility.} We have publicized our codes at https://github.com/weikai-li/HierarchicalMoE. The readme file contains instructions on how to run the codes. We have introduced the important hyper-parameters in this appendix, while other hyper-parameters can be found in the ``src/config.py,'' We use NVIDIA L40S (40GB) to train our model. It takes about 5 hours to train the model on the source kernels and about 1 hour to fine-tune it on the target kernels. Training costs about 30 GB of memory and fine-tuning costs about 12 GB of memory. We load the whole dataset into the memory before training. If the memory is not sufficient, we have an argument to disable this feature conveniently, and the memory cost for training can be reduced to about 15 GB. We run HLS on an AMD EPYC 7V13 64-core processor. We run HLS for the selected top-10 design points in parallel. It takes about 10 CPUs, 50 GB of memory, and 3 hours to run 10 design points for one kernel.

\section{Limitations and Future Work}

Though we have made much progress on domain generalization for HLS prediction, domain generalization is still a big challenge for HLS prediction models. First, the hierarchical MoE model does not have a huge advantage over HARP on every kernel. As shown in Table~\ref{table:main_results} in the main paper, it is significantly better than HARP on ``Fd'' and ``Gemv,'' slightly better on ``Gemm'' and ``Tr,'' but it is the same as HARP or worse than HARP on ``Sy'' and ``Ja.'' This might be due to the inherent difficulty of this problem. Future research is needed to develop even stronger models.

Second, HLS prediction is very difficult on complex kernels. In our experiments of domain transfer on complex kernels, the MSE is much larger than on the HLSyn kernels, as shown in Table~\ref{table:complex_kernels}. Although hierarchical MoE performs better than the baseline models, its MSE is still quite large. It demonstrates that hierarchical MoE is useful in improving the domain generalization ability, but it remains a future work to generally improve the model's ability of HLS prediction on complex kernels.

Besides, the hierarchical MoE structure is mainly useful in quick domain adaption using a small amount of data to fine-tune. When the computing resource is abundant enough to run a lot of HLS on the target kernels, we will have more data points for tine-tuning, and in this case, the hierarchical MoE model would perform similarly to HARP. The proposed model should be used in suitable applications.

\end{document}